\documentclass{article}

\usepackage{microtype}
\usepackage{graphicx}
\usepackage{subfigure}
\usepackage{booktabs} %
\usepackage{multirow}
\usepackage{hyperref}
\usepackage{xcolor}

\usepackage[accepted]{icml2025}

\usepackage{amsmath}
\usepackage{amssymb}
\usepackage{mathtools}
\usepackage{amsthm}

\usepackage[capitalize,noabbrev]{cleveref}

\newcommand{\pangu}{openPangu}
\newcommand{\eg}{\textit{e.g.}}
\newcommand{\modelname}{\pangu~Embedded-1B}
\theoremstyle{plain}

\theoremstyle{definition}

\theoremstyle{remark}

\usepackage[textsize=tiny]{todonotes}
\usepackage[hang,flushmargin]{footmisc}
\begin{document}

\twocolumn[

\icmltitle{Revealing the Power of Post-Training for Small Language Models via Knowledge Distillation}
\icmlsetsymbol{equal}{\dag}
\icmlsetsymbol{cc}{\#}
\icmlsetsymbol{pl}{*}
\begin{icmlauthorlist}
\icmlauthor{Miao Rang}{equal}
\icmlauthor{Zhenni Bi}{equal}
\icmlauthor{Hang Zhou}{}
\icmlauthor{Hanting Chen}{}
\icmlauthor{An Xiao}{}
\icmlauthor{Tianyu Guo}{}
\icmlauthor{Kai Han}{pl}
\icmlauthor{Xinghao Chen}{cc}
\icmlauthor{Yunhe Wang}{cc}

\end{icmlauthorlist}
\begin{center}
    Huawei Noah's Ark Lab
\end{center}
\icmlcorrespondingauthor{Xinghao Chen}{xinghao.chen@huawei.com}
\icmlcorrespondingauthor{Yunhe Wang}{yunhe.wang@huawei.com}

\icmlkeywords{Machine Learning, ICML}

\vskip 0.3in
]

{\renewcommand{\thefootnote}{}
\footnotetext{%
    \dag Equal contribution. \ 
    *Project Leader. \ 
    \#Corresponding Author. \ 
     \newline 
     \raggedright
    Correspondence to: 
    \newline 
    Xinghao Chen \textless xinghao.chen@huawei.com\textgreater , 
    \newline 
    Yunhe Wang \textless yunhe.wang@huawei.com\textgreater.
}
}

\begin{abstract}

The rapid advancement of large language models (LLMs) has significantly advanced the capabilities of artificial intelligence across various domains. However, their massive scale and high computational costs render them unsuitable for direct deployment in resource-constrained edge environments. This creates a critical need for high-performance small models that can operate efficiently at the edge. Yet, after pre-training alone, these smaller models often fail to meet the performance requirements of complex tasks. To bridge this gap, we introduce a systematic post-training pipeline that efficiently enhances small model accuracy. Our post training pipeline consists of curriculum-based supervised fine-tuning (SFT) and offline on-policy knowledge distillation. The resulting instruction-tuned model achieves state-of-the-art performance among billion-parameter models, demonstrating strong generalization under strict hardware constraints while maintaining competitive accuracy across a variety of tasks. This work provides a practical and efficient solution for developing high-performance language models on Ascend edge devices.

\end{abstract}

\section{Introduction}
Large Language Models (LLMs) have transformed the landscape of artificial intelligence by leveraging self-attention mechanisms and massive-scale pre-training to capture hierarchical linguistic patterns, semantic relationships, and cross-domain knowledge~\cite{achiam2023gpt}. Open-sourced models such as LLaMA~\cite{touvron2023llama}, DeepSeek~\cite{liu2024deepseek}, Qwen~\cite{bai2023qwen,yang2025qwen3} and openPangu~\cite{chen2025pangu,tang2025pangu} have demonstrated exceptional performance in complex tasks including text generation, reasoning, and multilingual understanding. However, their success hinges on enormous parameter counts (\eg, DeepSeek-V3~\cite{liu2024deepseek} has 671B parameters in total) and extensive computational resources, which limit their deployment in latency-sensitive or resource-constrained environments. The prohibitive energy consumption and hardware requirements of LLMs have raised critical challenges for real-world applications, particularly in scenarios demanding on-device processing, privacy preservation, or low-latency responsiveness.

To address these limitations, a paradigm shift has emerged toward developing efficient Small Language Models (SLMs) for edge devices and resource-constrained platforms. The training pipeline for small models is divided into pre-training and post-training. As is well known, pre-training focuses on building foundational capabilities and requires substantial data and training resources. In contrast, post-training concentrates on enhancing these abilities and typically demands significantly fewer resources. Therefore, leveraging post-training to substantially improve the accuracy of small models under resource constraints has become a critical area of research. Existing post-training methods include supervised fine-tuning~\cite{lobo2024impact, luong2024reft}, and knowledge distillation (e.g., GKD~\cite{agarwal2024policy}).

Building upon these established techniques, our work introduces a multi-stage post-training pipeline designed to enhance small language models(e.g., \modelname\footnote{https://ai.gitcode.com/ascend-tribe/openPangu-Embedded-1B-V1.1} which is specifically developed for efficient inference on Ascend edge devices). Our process begins with SFT using a curated curriculum that transitions from step-by-step reasoning to fast-response examples, thereby strengthening the model's fundamental instruction-following abilities. The instruct model after SFT is named as \modelname-SFT.

Subsequently, we employ offline on-policy knowledge distillation from a larger teacher model that shares the same tokenizer, enabling efficient logit-level knowledge transfer. The resulting instruction-tuned model, named \modelname-KD, benefits from this comprehensive 
post-training pipeline and achieves state-of-the-art accuracy among billion-scale instruction models.

\section{Related Work}
\textbf{Small Language Models.} 
The development of small language models in the 1 billion parameter range has progressed through several distinct phases, evolving from initial capability demonstrations to highly optimized systems. 
Initial explorations, such as OPT-1.3\,B~\cite{zhang2022opt} and GPT-Neo-1.3\,B~\cite{kashyap2022gpt}, first established this scale as an efficient sweet spot for natural language understanding (NLU) tasks. 
Building on this foundation, subsequent research shifted its focus toward the critical role of data quality, with models like  DeepSeek-Coder-1.3\,B~\cite{guo2024deepseek} demonstrating that curated, domain-specific corpora could enable smaller models to surpass much larger baselines on expert tasks. 
This data-centric paradigm is best exemplified by Phi-1/1.5~\cite{fu2024hardware}, which achieve remarkable performance on HumanEval using a small, ``textbook-quality'' dataset, achieving over 50\,\% pass@1.
The focus then shift to architecture-algorithm co-design in 2024, with innovations like MobileLLM~\cite{liu2024mobilellm} introducing latency-efficient architectures, TinyLlama-1.1\,B~\cite{zhang2024tinyllama} showcasing the benefits of extreme-scale training on 3 trillion tokens, and the fully open-source OLMo-1\,B~\cite{groeneveld2024olmo} becoming a standard testbed for scaling-law research. 

\textbf{Model-aware Training.}
Earlier work in LLM post-training primarily focuses on selecting high-quality training data from a general perspective without considering model-specific issues~\cite{liumakes,zhou_lima_2023}. While this approach has been effective in improving model performance, recent studies have shown that data distributions significantly deviating from the base model’s can be difficult for the model to learn from and may even degrade performance~\cite{ren_learning_2024}. As a result, more research advocates for model-specific data selection~\cite{du2023mods,li-etal-2024-quantity}. These studies do not explore distillation scenarios, and our work further extends this concept by integrating an iterative distillation pipeline, leveraging our proposed model-aware complexity score.

\textbf{Knowledge Distillation.} Knowledge Distillation (KD) is a widely adopted model compression technique where a compact student model is trained to replicate the behavior of a larger teacher model~\cite{hinton2015distilling}, facilitating the deployment of high-performance models in resource-constrained settings. A fundamental challenge in applying KD to autoregressive models is the exposure bias phenomenon~\cite{ross2011reduction}—a discrepancy between the training distribution, which is conditioned on ground-truth sequences, and the inference distribution, where the model conditions on its own generated outputs. This mismatch can lead to an accumulation of errors during generation. To directly mitigate this issue, on-policy distillation methods have been introduced. These approaches train the student on sequences it generates dynamically, with the teacher providing supervision on these self-generated samples~\cite{agarwal2024policy}. While this methodology effectively aligns the training and inference distributions, its iterative nature—requiring repeated generation and training steps—incurs substantial computational overhead. To harness the benefits of on-policy training while maintaining offline efficiency, our work introduces a novel offline adaptation of this principle. Our framework is executed in a straightforward two-stage process: first, the student model conducts a single inference pass over the training corpus to generate a complete set of responses. Second, conventional logits-based distillation is performed using this student-generated dataset, with the teacher model providing the supervisory soft labels. This method effectively simulates an on-policy data distribution, thereby alleviating exposure bias while circumventing the costly iterative loop inherent to true on-policy methods. The result is a simple and scalable framework that markedly improves training efficiency over online counterparts without a significant compromise in model performance.

\section{Post-training Strategy}
\begin{figure*}[htbp]
    \centering 
    \includegraphics[width=0.8\textwidth]{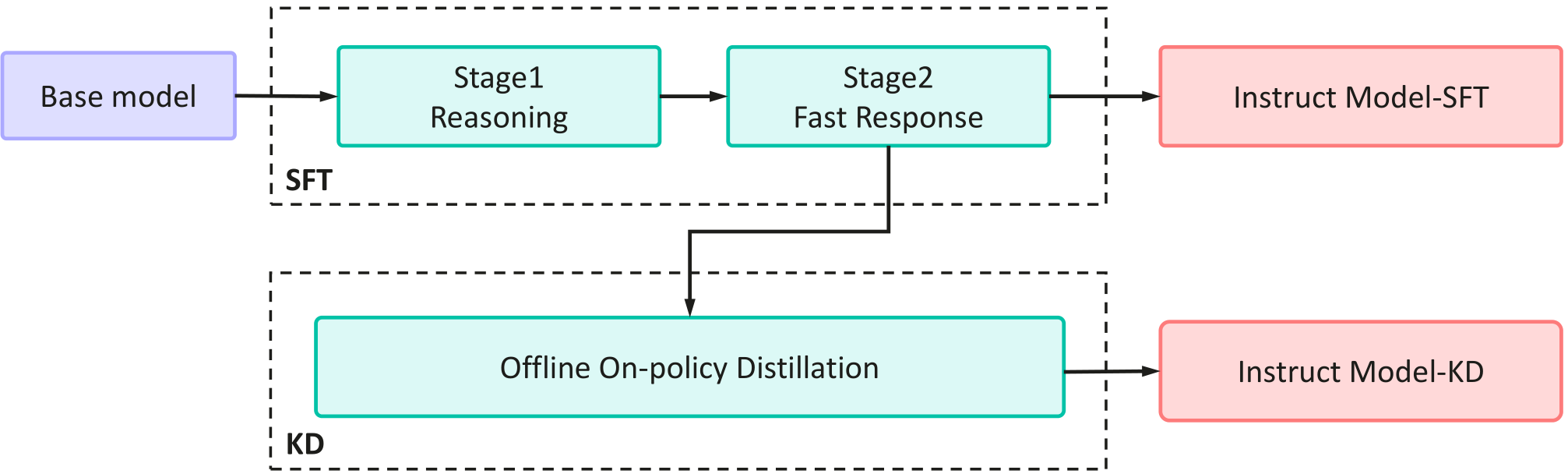} 

    \caption{An illustration of the \modelname~post-training pipeline. The pipeline consists of two primary stages:
Two-Stage Curriculum SFT and Offline On-policy Knowledge Distillation.}
    \label{fig:pipeline} 
\end{figure*}

In this section, we present the post-training strategy for our base model \modelname{}. This strategy includes Two-Stage Curriculum SFT, and Offline On-policy Knowledge Distillation, as shown in Fig~\ref{fig:pipeline}.

\subsection{Post-training Data}
For post-training data, core principles of high quality, diversity, and complexity are prioritized, with a reasoning-centered design to enhance the model's capabilities and generalization as well. The initial data pool integrates multi-source data: open-source instruction datasets, real-world industrial queries (\textit{e.g.,} finance/healthcare scenarios), and synthetic problems derived from pre-training corpora. The post-training data are split into two key subsets: reasoning tasks (advanced STEM with multi-step computation, code generation requiring symbolic manipulation, logical inference) and non-reasoning tasks (general QA, text analysis, long-context understanding, semantic classification, tool use, and agent), with a 3:1 sampling ratio adopted. 

To ensure data quality, a rigorous two-stage processing pipeline is constructed: 1) Prior filtering: we leverage our models to annotate data with attributes (subcategory, question type, answer verifiability, difficulty metrics like reasoning hops) to filter unqualified samples; 2) Diversity maintenance: We use N-gram-based MinHash-LSH to eliminate near-duplicates, followed by the ZIP algorithm (guided by entropy~\cite{yin2024entropylawstorydata}) to select samples by prioritizing low compression ratios (higher diversity) and minimizing similarity to existing entries, to yield a pattern-rich dataset. The overall post-training dataset emphasizes reasoning tasks as well as dataset diversity, thus strengthening the model's ability to avoid superficial pattern matching, while non-reasoning tasks support basic language competence, collectively enhancing performance across both specialized reasoning tasks and general abilities.

\subsection{Two-Stage Curriculum SFT}

During the SFT phase, we employ Two-Stage Curriculum SFT~\cite{chen2025pangu}, a method grounded in cognitive science: robust reasoning skills must first be explicitly learned before they can be applied intuitively. In Stage 1, the model is trained on reasoning-enhanced data containing explicit, step-by-step chains of thought to build strong inferential capabilities and avoid shallow heuristics. In Stage 2, training shifts to standard prompt–response pairs without intermediate reasoning steps, encouraging the model to implicitly apply its learned reasoning framework to produce concise, accurate outputs. This approach effectively enhances reasoning ability, overall performance, and response efficiency.

\subsection{Offline On-policy Knowledge Distillation}
On-policy distillation~\cite{agarwal2024policy,lin2020autoregressive} has been proven to be an effective model compression technique. It leverages the outputs of a larger, more powerful teacher model to guide a smaller student model, thereby enhancing the student's accuracy and generalization capabilities. This process enables compact student models to achieve superior performance.

To decouple the generation of teacher model guidance data from the training process of the student model, and to allow independent optimization of the training data and process, we design an \textit{offline version of on-policy distillation}. Unlike the online version, which cannot dynamically adjust and optimize the quality of the generated logits, the offline version allows for selecting high-quality data during the preprocessing stage to generate teacher model guidance information. Compared to the standard SFT process, our method introduces only an additional Knowledge Distillation (KD) loss term. This design simplifies the implementation process, ensuring ease of use and straightforward operation, while enhancing the flexibility and controllability of the system.

Our method is systematically structured into two distinct phases: an offline data preparation phase followed by a training phase. The offline data preparation phase includes two key components: Student-driven Response Generation and Teacher's Token-Level Logits Prediction:

\begin{itemize}%

\item \textbf{Student-driven Response Generation:} The process commences with our SFT-trained student model, which performs inference on the queries from the original training dataset. This generates an intermediate on-policy dataset, $D_s$, where each sample consists of a query and the corresponding student-generated response. This initial step, which we refer to as \textit{distillation by student}, ensures that the subsequent teacher guidance is grounded in the student's own output distribution.

\item \textbf{Teacher's Token-Level Logits Prediction:} Following response generation, we employ a powerful teacher model(e.g., a 7B parameter model sharing the same tokenizer) to annotate each sequence in $D_s$ with token-level guidance. For each position $n$ in a student-generated sequence, the teacher model is conditioned on the prefix of the preceding $n-1$ tokens to predict the distribution for the $n$-th token. Crucially, rather than performing greedy decoding, we preserve the teacher's full predictive distribution by recording the logits for the top-k most probable tokens. This constrained conditioning strategy is the cornerstone of our approach. By compelling the teacher to generate guidance from a probabilistic space already accessible to the student, we effectively minimize the intrinsic distributional divergence between the two models. This facilitates a more stable and efficient knowledge transfer.

\end{itemize}
\paragraph{Training with a Composite Loss Function.} In the training phase, the student model is optimized using a composite loss function that synergistically combines standard supervised learning with knowledge distillation. The total loss $L_{total}$ is formulated as the weighted sum of the Cross-Entropy (CE) loss $L_{CE}$ and the knowledge distillation loss $L_{KD}$:
$$L_{total} = (1 - \lambda_{KD}) \cdot L_{CE} + \lambda_{KD} \cdot L_{KD}$$
where $\lambda_{KD}$ is a scalar hyperparameter that is a weighting coefficient. It meticulously balances the influence of the direct supervised objective ($L_{CE}$) and the teacher's distributional guidance ($L_{KD}$).
The core of the distillation process is the minimization of the KL divergence, denoted $D_{KL}(P \parallel Q)$, is an asymmetric measure that quantifies how a probability distribution $Q$ (from the student) differs from a reference probability distribution $P$ (from the teacher).
For discrete distributions over a vocabulary of classes C, it is defined as:
\[
D_{KL}(P \parallel Q) = \sum_{c \in C} P(c) \log \frac{P(c)}{Q(c)}
\]
By minimizing the KL divergence, the student's output distribution ($Q$) is trained to approximate the soft-target distribution ($P$) provided by the teacher. A lower divergence value signifies that the student has successfully learned to mimic the teacher's predictive patterns, effectively internalizing the knowledge transferred during distillation.

\section{Post-training Evaluation}
 
\subsection{Main Results}
In this section, we evaluate \modelname{}-KD, which is trained using our post-training techniques, on both reasoning and normal language tasks.
\begin{table*}[t]
    \centering
    \footnotesize
    \setlength{\tabcolsep}{4.5pt}
    \caption{Instruct model (non-thinking) comparison between \modelname{}-KD and other representative models across a diverse set of benchmarks for evaluating language and reasoning skills. \textbf{Bold} values represent the best results in each line among models at the 1B-parameter scale. If the original paper reports the results, we present the results from the original paper (marked with asterisks $^*$); otherwise, we list our reproduced results.}
    \resizebox{1.0\textwidth}{!}{
    \small
    \begin{tabular}{@{}c l | c c| c  c  c c| c c}
    \toprule
    & \multirow{2}{*}{\centering \textbf{Benchmark {\tiny (Metric)}}}   & \multirow{2}{*}{\centering\textcolor{black!60}{\textbf{Qwen3}}}   &\multirow{2}{*}{\centering\textcolor{black!60}{\textbf{Qwen2.5}}}   & \multirow{2}{*}{\centering\textbf{Gemma3}} & \multirow{2}{*}{\centering\textbf{Llama3.2}} & \multirow{2}{*}{\centering{\textbf{Qwen3}}} & \multirow{2}{*}{\centering\textbf{MiniCPM4}} &  \textbf{\pangu~Embedded} &  \textbf{\pangu~Embedded}\\
    &   &   &   &   & &  &  & SFT & KD \\
    \midrule
    & \# Total Params  & \textcolor{black!60}{1.7B} & \textcolor{black!60}{1.5B} & 1B & 1B &0.6B & 0.5B &  1B &  1B \\
    \midrule
    \multirow{5}{*}{\textbf{General}} & MMLU{\tiny (Acc)} &    \textcolor{black!60}{63.37} & \textcolor{black!60}{52.84}  & 37.49 & 32.19  & 44.24 & 55.55$^*$ &  63.21 &  \textbf{67.28} \\
    & CMMLU {\tiny (Acc)}& \textcolor{black!60}{61.22}  & \textcolor{black!60}{53.98}   & 31.57 & 10.81 & 42.94 & \textbf{65.22}$^*$ &  53.10 & 57.75  \\
    
    & C-Eval {\tiny (Acc)}&  \textcolor{black!60}{61.00$^*$} & \textcolor{black!60}{59.30} & 32.49  & 32.08& 42.60$^*$  & 66.11$^*$ &  58.51 & \textbf{66.55} \\
    
    & IF-Eval {\tiny (Prompt Strict)} & \textcolor{black!60}{68.20$^*$} & \textcolor{black!60}{42.50$^*$} & 51.57  & 39.37 &54.50$^*$  & 50.28  & 56.38 & \textbf{62.66}  \\
    
    & CLUEWSC{\tiny (Acc)}  & \textcolor{black!60}{77.36} & \textcolor{black!60}{74.59} &  50.20 & 52.36 & 50.31 & 49.90 &  76.95 & \textbf{80.02} \\
    
    \midrule
    \multirow{2}{*}{\textbf{Math}} & GSM8K {\tiny (Acc)} & \textcolor{black!60}{77.03} & \textcolor{black!60}{73.20$^*$}  & 57.16 & 36.39 &  59.29 & 52.08$^*$ & 70.89 & \textbf{77.33} \\ 
    & MATH-500 {\tiny (Acc)} & \textcolor{black!60}{73.00$^*$} & \textcolor{black!60}{46.60}  & 39.40  & 18.20 & 55.20$^*$ & 29.60$^*$ & 56.20 & \textbf{73.80} \\
    \midrule
    \multirow{2}{*}{\textbf{Reasoning}}& DROP {\tiny (F1)}  & \textcolor{black!60}{61.21} & \textcolor{black!60}{48.44} & 30.98  & \textbf{45.23} &  34.69 & 30.07 & 28.73 & 40.95  \\ 
    & GPQA-Diamond {\tiny (Pass@1)}   & \textcolor{black!60}{28.60$^*$} & \textcolor{black!60}{29.80$^*$}  & 19.20$^*$ & 29.29 & 22.90$^*$ & 28.28 & 43.43 & \textbf{44.44} \\
    \midrule
    \multirow{2}{*}{\textbf{Code}} & MBPP {\tiny (Pass@1)}   & \textcolor{black!60}{60.70} & \textcolor{black!60}{63.20$^*$}  & 58.75  & 40.08 &  46.69 & 59.14$^*$ & 52.53 & \textbf{61.09}  \\
    & HumanEval {\tiny (Pass@1)}  & \textcolor{black!60}{68.90} & \textcolor{black!60}{61.60$^*$} & 40.24  & 29.88  & 40.85 & 46.34$^*$ & 59.15 & \textbf{65.85} \\
    \midrule
    & Average & \textcolor{black!60}{63.69} & \textcolor{black!60}{55.10}   & 40.82 & 33.26 & 44.93 & 48.42 & 56.28 & \textbf{63.43} \\
    \bottomrule
    \end{tabular}
    }
    \label{tab:benchmark-post-training}
\end{table*}

\paragraph{SFT Training Setup.}
The SFT phase of \modelname~uses a two-stage fine-tuning approach, each stage running for 10 epochs. For overall training stability, we employ the AdamW optimizer with a weight decay of 0.1 and apply gradient clipping at a threshold of 1.0. We set the maximum sequence length to 32K to maximize computational efficiency, packing multiple samples into each sequence.

The first stage focuses on complex reasoning, using a global batch size of 4 million tokens. Its learning rate follows a cosine schedule with a 200-iteration warmup, annealing from a peak of $2\times 10^{-5}$ down to $2\times 10^{-6}$. The second stage targets open-ended generation tasks, using a smaller global batch size of 2 million tokens. The learning rate for this stage also follows a cosine schedule, decaying from $1\times 10^{-5}$ to $1\times 10^{-6}$.

\paragraph{Baselines \& Benchmarks.}  The \modelname~family is compared against several prominent open-source models within a similar parameter class to ensure a relevant and competitive analysis. The compared baselines include Qwen3 (1.7B and 0.6B)~\cite{yang2025qwen3}, Qwen2.5(1.5B)~\cite{Yang2024Qwen25TR}, Gemma3 (1B)~\cite{team2024gemma}, Llama3.2 (1B)~\cite{dubey2024llama} and  MiniCPM4 (0.5B)~\cite{team2025minicpm4}. Including the larger Qwen3-1.7B model serves as a critical point of comparison, allowing for an evaluation of the parameter efficiency of the proposed approach. Performance is measured using standard metrics appropriate for each benchmark. Accuracy (Acc) is used for tasks with single correct answers, such as MMLU~\cite{hendrycks2020measuring} and GSM8K~\cite{cobbe2021training}. The F1 Score is employed for tasks like DROP~\cite{Dua2019DROPAR}, which require a balance of precision and recall in text extraction. For code generation tasks like MBPP~\cite{austin2021program} and HumanEval~\cite{Chen2021EvaluatingLL}, Pass@1 is used, which measures the percentage of problems for which a correct solution is generated in a single attempt.

To ensure a fair and comprehensive evaluation,  the evaluation suite is meticulously curated to probe a wide range of cognitive abilities, categorized into four primary domains:
\begin{itemize}%
\item \emph{General tasks}: Fundamental language understanding, multi-domain knowledge, and chinese language proficiency are assessed using established benchmarks. These include MMLU for broad, multi-disciplinary knowledge; CMMLU~\cite{Li2023cmmlu} and C-Eval~\cite{Huang2023CEvalAM} for comprehensive Chinese language evaluation; IF-Eval~\cite{zhou2023instruction} for instruction-following fidelity; and CLUEWSC~\cite{xu2020clue} for commonsense reasoning. These benchmarks test a model's core knowledge base and ability to apply it in varied contexts.

\item \emph{Mathematics}: To evaluate complex, multi-step quantitative reasoning, the evaluation employs GSM8K and MATH-500~\cite{hendrycks2021measuring}. These benchmarks require numerical computation and the critical ability to translate natural language problems into logical steps, testing the depth of a model's reasoning capacity.

\item \emph{Reasoning}: The models' capacity for complex reasoning and information extraction is tested using DROP~\cite{Dua2019DROPAR}, which measures reading comprehension intertwined with arithmetic reasoning, and GPQA-Diamond~\cite{rein2024gpqa}. This question-answering dataset that probes deep, domain-specific reasoning in physics and chemistry.

\item \emph{Code Generation}: Proficiency in programming is evaluated using MBPP~\cite{austin2021program} and HumanEval~\cite{Chen2021EvaluatingLL}. These benchmarks assess the ability to synthesize correct and functional code from natural language docstrings, a key skill for practical applications.

\end{itemize}

\paragraph{Evaluation Results.}

The empirical results (summarized in Table~\ref{tab:benchmark-post-training}) unequivocally establish the superior performance of the \pangu~Embedded model family, with the \modelname-KD setting a new state-of-the-art for models in the 1B parameter class. A salient finding is its aggregate performance, achieving an average score of 63.43, which is on par with the larger Qwen3-1.7B model (63.69). This demonstrates exceptional parameter efficiency, suggesting that advanced training and alignment methodologies can be more impactful than simply scaling model size. The model's most profound advantage lies in mathematical and complex reasoning, where it achieves leading scores on GSM8K (77.33) and MATH-500 (73.80). This strength is complemented by robust general knowledge capabilities, securing top positions on benchmarks like CLUEWSC (80.02) and MMLU (67.28), highlighting a robust and well-rounded bilingual foundation.

\begin{table*}[htbp]
    \centering
    \caption{Accuracy comparison of SFT methods with greedy decoding. The Reasoning (w/ CoT) to Fast approach can effectively improve accuracy during the SFT stage.}
    \label{tab:sft_benchmark_category}
    \resizebox{\linewidth}{!}{%
    \begin{tabular}{l|ccccc|cc|cc|cc|c}
        \toprule
        \multirow{3}{*}{\textbf{Method}} & \multicolumn{5}{c|}{\textbf{General}}
        & \multicolumn{2}{c|}{\textbf{Math}} 
        & \multicolumn{2}{c|}{\textbf{Reasoning}}
        & \multicolumn{2}{c|}{\textbf{Code}} 
        & \multirow{3}{*}{\textbf{AVG}} \\
        \cmidrule(lr){2-6} \cmidrule(lr){7-10} \cmidrule(lr){11-12}
        
         & \textbf{MMLU} & \textbf{CMMLU} & \textbf{C-Eval} & \textbf{IF-Eval} & \textbf{CLUEWSC} & \textbf{GSM8K} & \textbf{MATH-500} & \textbf{DROP} & \textbf{GPQA-Diamond} & \textbf{MBPP} & \textbf{HumanEval} &  \\
        \midrule
        
        Direct Fast-Response & 60.83 & \textbf{49.58} & 60.06 & 56.19 & 71.21 & 70.20 & 46.60 & 55.84 & 32.83 & 54.47 & 54.88 & 55.70 \\
        Reasoning (w/o CoT) $\rightarrow$ Fast & 61.76 & 52.02 & 62.66 & 61.18 & 74.18 & 70.36 & 51.60 & 42.60 & 34.85 & 54.86 & 56.71 & \textbf{56.62} \\
        \textbf{Reasoning (w/ CoT) $\rightarrow$ Fast} & 63.21 & 53.10 & \textbf{58.51} & 56.38 & 76.95 & \textbf{70.89} & 56.20 & 28.73 & \textbf{43.43} & \textbf{52.53} & \textbf{59.15}  & 56.28 \\
        \bottomrule
    \end{tabular}%
    }
\end{table*}

\subsection{SFT}
To optimize the SFT process, we conduct a series of ablation studies to investigate whether a multi-stage training curriculum can enhance the model's "fast thinking" or intuitive response capabilities. Specifically, we seek to determine whether first fine-tuning on deliberative reasoning tasks before fine-tuning on rapid-response data yields superior performance. We evaluate three distinct training methods:

\begin{itemize}
    \item \textbf{One-Stage Curriculum (Direct Fast Response).} In this single-stage approach, the model is directly fine-tuned using only the fast response dataset. This baseline measures the efficacy of training exclusively on target-domain data without any preparatory learning phases.

    \item \textbf{Two-Stage Curriculum (Reasoning-to-Fast, without CoT).} This method includes a two-stage curriculum. The model is first trained on a Reasoning dataset from which the intermediate reasoning steps (i.e., the "thought process") have been explicitly removed. Following this, the model is fine-tuned on the fast response dataset. This approach tests the benefit of a sequential training regimen on datasets with different characteristics, without explicitly teaching the model to reason.

    \item \textbf{Two-Stage Curriculum (Reasoning-to-Fast, with CoT).} \modelname~ is also trained in two stages. However, in the initial stage, it is fine-tuned on the complete Reasoning dataset, which crucially retains the detailed reasoning chains and thought processes. This stage is designed to instill deliberative reasoning capabilities before fine-tuning for rapid-response generation.

\end{itemize}

The experimental outcomes highlight the efficacy of our proposed two-stage curriculum, as both the ``Reasoning (w/ CoT)-to-Fast" and ``Reasoning (w/o CoT)-to-Fast" approaches significantly outperform the direct ``Fast-Response" fine-tuning method. Notably, the two curriculum variants perform comparably—with the non-CoT approach achieving a peak accuracy of 56.62$\%$—indicating that the curriculum structure itself is the primary driver of success, even without explicit reasoning paths. This initial reasoning phase acts as a form of cognitive scaffolding, equipping the model with foundational problem-solving skills. dur the second fast response phase, the model is not merely memorizing input-output pairs but can leverage its acquired reasoning abilities to generate more robust and accurate responses.  These findings demonstrate the profound value of a curriculum that prioritizes the development of underlying skills before optimising for rapid task completion.

\subsection{Knowledge Distillation}
We conduct a systematic ablation study on knowledge distillation to further enhance model performance. Our analysis focuses on four key dimensions: (i) the weighting of the distillation loss term, (ii) the effect of the top-k value during decoding, (iii) the choice of distillation strategy. This investigation aims to elucidate each factor's relative importance and identify the most effective configuration.

\subparagraph{Knowledge Distillation Loss Weight.}
To optimize the balance between the standard cross-entropy loss and the distillation loss, we conduct experiments with different values of the KD loss weight ($\lambda_{KD}$), ranging from 0.5 to 1.0. To accelerate the experimental cycle, we conduct distillation experiments based on a single stage (Direct Fast Response) of SFT. This approach introduces a new Stage~2 to perform distillation guided by labels, with the distillation learning rate kept consistent with that used in SFT. This systematic comparison shows setting $\lambda_{KD}=0.9$ yields the best overall performance, as shown in Table~\ref{tab:distillation_loss_weight}.

\begin{table}[h]
    \centering
    \caption{Effect of KD loss weight on the average benchmark performance of stage2 distillation by label. 
    The evaluation metric is the average zero-shot accuracy across eight benchmarks: 
    MMLU, CMMLU, CEval, BBH, GSM8K, MATH, MBPP, and HumanEval. All models use greedy decoding with a decode length of 8K.}
    \label{tab:distillation_loss_weight}
    \resizebox{\linewidth}{!}{%
    \begin{tabular}{lcccccc}
        \toprule
        $\lambda_{KD}$ & 0.5 & 0.6 & 0.7 & 0.8 & \textbf{0.9} & 1.0 \\
        \midrule
        Average  & 53.94 & 54.07 & 53.88 & 54.43 & \textbf{55.29} & 54.44 \\
        \bottomrule
    \end{tabular}
    }
\end{table}

\subparagraph{Top-$k$ Value Effect.}
To explore the impact of the number of top tokens used in the distillation process, we experiment with four values for the top-$k$ parameter: 5, 10, 15 and 20. By adjusting the value of $k$, we aim to investigate how the number of candidate logits influences model performance. The experimental setup is consistent with that described in the \emph{Knowledge Distillation Loss Weight} section. We observe that training time remains almost identical across different $k$ values. In terms of performance, increasing $k$ from 5 to 10 yields an improvement of 0.51$\%$, with top-$k$=10 achieving the highest average accuracy. However, further enlarging $k$ to 15 or 20 leads to a degradation in performance, as shown in Table~\ref{tab:topk_value}. Hence, we adopt $\lambda_{KD}=0.9$ and top-$k$=10 by default unless otherwise specified.

\begin{table}[h]
    \centering
    \caption{Effect of top-$k$ value on the average benchmark performance of stage2 distillation by label. 
    The evaluation metric is the average zero-shot accuracy across eight benchmarks: 
    MMLU, CMMLU, CEval, BBH, GSM8K, MATH, MBPP, and HumanEval. All models use greedy decoding with a decode length of 8K, and the KD loss weight is fixed at $\lambda_{KD}=0.9$.}
    \label{tab:topk_value}
    \begin{tabular}{lcccc}
        \toprule
        Top-$k$ & 5 & \textbf{10} & 15 & 20 \\
        \midrule
        Average Performance & 54.78 & \textbf{55.29} & 54.24 & 54.40 \\
        \bottomrule
    \end{tabular}
\end{table}

\begin{table*}[htbp]
    \centering
    \caption{Ablation study on knowledge distillation. We compare three approaches: "Distillation by Label" where teacher logits are conditioned on ground-truth labels and guided by its top-10 predictions; "Distillation by Teacher" where logits are conditioned on the teacher's own generated response; and "Distillation by Student" where logits are conditioned on the student's predictions to encourage self-consistency. The asterisk (*) denotes that the student model is updated twice with its latest parameters during training for response generation. All results are based on greedy decoding with a generation length of 8K tokens.}

    \label{tab:distillation_timing}
    \resizebox{\linewidth}{!}{%
    \begin{tabular}{l|ccccc|cc|cc|cc|c}
        \toprule
        \multirow{3}{*}{\textbf{Method}} & \multicolumn{5}{c|}{\textbf{General}}
        & \multicolumn{2}{c|}{\textbf{Math}} 
        & \multicolumn{2}{c|}{\textbf{Reasoning}}
        & \multicolumn{2}{c|}{\textbf{Code}} 
        & \multirow{3}{*}{\textbf{AVG}} \\
        \cmidrule(lr){2-6} \cmidrule(lr){7-10} \cmidrule(lr){11-12}
        
         & \textbf{MMLU} & \textbf{CMMLU} & \textbf{C-Eval} & \textbf{IF-Eval} & \textbf{CLUEWSC} & \textbf{GSM8K} & \textbf{MATH-500} & \textbf{DROP} & \textbf{GPQA-Diamond} & \textbf{MBPP} & \textbf{HumanEval} &  \\
        \midrule
        
        Stage2 SFT & 63.21 & 53.10 & 58.51 & 56.38 & 76.95 & 70.89 & 56.20 & 28.73 & 43.43 & 52.53 & 59.15 & 56.28 \\ 
        Stage3 Distillation by Label & 62.68 & 53.97 & 58.83 & 59.70 & 77.56 & 72.10 & 65.20 & 30.24 & 46.46 & 57.20 & 62.80 & 58.79  \\ 
        Stage3 Distillation by Teacher & 61.74 & 54.94 & 62.83 & 59.89 & 79.92 & 74.30 & 61.20 & 36.59 & \textbf{47.98} & 55.25 & 60.37 & 59.55  \\
        \textbf{Stage3 Distillation by Student} & 65.91 & 56.17 & 64.30 & 59.33 & 79.51 & 76.72 & 71.40 & \textbf{40.99} & 44.95 & 58.75 & \textbf{67.07} & 62.28  \\ 
        \textbf{Stage3 Distillation by Student*} & \textbf{67.28} & \textbf{57.75} & \textbf{66.55} & \textbf{62.66} & \textbf{80.02} & \textbf{77.33} & \textbf{73.80} & 40.95 & 44.44 & \textbf{61.09} & 65.85 & \textbf{63.43}  \\ 
        \bottomrule
    \end{tabular}%
    }
\end{table*}

\begin{table*}[htbp]
    \centering
    \caption{Ablation study on post-train pipeline. KD use distillation by label which teacher logits are conditioned on ground-truth labels, with the teacher's topk-10 predicted logits serving as auxiliary guidance. All models use greedy decoding with a decode length of 8K.}

    \label{tab:post_train_ablation}
    \resizebox{\linewidth}{!}{%
    \begin{tabular}{l|ccccc|cc|cc|cc|c}
        \toprule
        \multirow{3}{*}{\textbf{Method}} & \multicolumn{5}{c|}{\textbf{General}}
        & \multicolumn{2}{c|}{\textbf{Math}} 
        & \multicolumn{2}{c|}{\textbf{Reasoning}}
        & \multicolumn{2}{c|}{\textbf{Code}} 
        & \multirow{3}{*}{\textbf{AVG}} \\
        \cmidrule(lr){2-6} \cmidrule(lr){7-10} \cmidrule(lr){11-12}
        
         & \textbf{MMLU} & \textbf{CMMLU} & \textbf{C-Eval} & \textbf{IF-Eval} & \textbf{CLUEWSC} & \textbf{GSM8K} & \textbf{MATH-500} & \textbf{DROP} & \textbf{GPQA-Diamond} & \textbf{MBPP} & \textbf{HumanEval} &  \\
        \midrule
        
        SFT(Reasoning) + SFT(Fast)  & \textbf{63.21} & 53.10 & 58.51 & 56.38 & 76.95 & 70.89 & 56.20 & 28.73 & 43.43 & 52.53 & 59.15 & 56.28 \\ 
        SFT(Reasoning) + KD(Fast) & 61.38 & \textbf{55.12} & 60.14 & 56.93 & 75.41 & 72.78 & 64.60 & 43.01 & 42.93 & 51.36 & 61.59 & 58.66 \\ 
        \midrule
        SFT(Reasoning) + SFT(Fast) + KD(Fast) & 62.68 & 53.97 & 58.83 & \textbf{59.70} & 77.56 & 72.10 & 65.20 & 30.24 & \textbf{46.46} & 57.20 & 62.80 & 58.79  \\
        SFT(Reasoning) + KD(Fast) + SFT(Fast) & 63.75 & 54.78 & \textbf{60.63} & 58.78 & 73.05 & 66.94 & 55.20 & \textbf{45.99} & 43.94 & 55.25 & 58.54 & 57.90  \\
        \midrule
        SFT(Reasoning) + SFT(Fast) + KD(Reasoning) + KD(Fast) & 62.69 & 53.95 & 62.24 & 57.86 & 79.61 & 73.09 & \textbf{66.20} & 33.61 & {39.39} & 56.42 & 60.37 & 58.68  \\
        SFT(Reasoning) + KD(Reasoning) + SFT(Fast) + KD(Fast) & 62.82 & 53.22 & 60.56 & 59.15 & \textbf{81.25} & \textbf{73.16} & 64.80 & 37.27 & {40.40} & \textbf{57.59} & \textbf{65.24} & \textbf{59.59}  \\
        \bottomrule
    \end{tabular}%
    }
\end{table*}

\subparagraph{Distillation Strategy.}
We conduct a key ablation study to compare three distinct distillation strategies. To simplify the experimental setup, we introduce an additional Knowledge Distillation (KD) stage to our two-stage Supervised Fine-Tuning (SFT) framework, specifically for distilling the 'fast-response' data. The first is the conventional approach, "Distillation by Label"~\cite{hinton2015distilling, sanh2019distilbert}, where teacher logits are conditioned on ground-truth labels. The second strategy, "Distillation by Teacher" ~\cite{kim2016sequence}, conditions the teacher's logits on its own generated response, which can better approximate the teacher model's intrinsic data distribution. The third strategy is an offline on-policy approach, "Distillation by Student", where the teacher provides target logits based on the student model's own generated response.

As shown in Table~\ref{tab:distillation_timing}, all methods prove effective. ``Distillation by Teacher" achieves a significant accuracy gain of 3.26$\%$. However, the results demonstrate the clear superiority of the "Distillation by Student" strategy, which achieves a more substantial performance increase of 6$\%$. This finding highlights the benefit of aligning the teacher's guidance with the student's current output space. By conditioning distillation on the student's response, the teacher provides corrective and refining signals on a distribution immediately relevant to the student's state. This alignment minimizes the distributional mismatch between the two models, creating a more stable learning signal and facilitating more effective knowledge transfer.

Furthermore, we find that the performance of the "Distillation by Student" strategy can be enhanced. By periodically updating the student model with its latest parameters during the training process—in our case, twice—and re-generating responses for subsequent distillation, the model's accuracy is further improved to 63.43$\%$. This creates a dynamic, self-correcting loop that maximizes the efficiency of knowledge transfer and significantly boosts the final performance of the student model.

\subsection{Ablation Study on Post-training Pipeline}
We further conduct an ablation study to investigate the optimal design of the post-training pipeline, focusing on how to interleave supervised fine-tuning and knowledge distillation. Here, KD is implemented as label-based distillation, where the teacher logits are conditioned on the ground-truth labels, and the teacher's top-$k$=10 predicted logits serve as auxiliary guidance. The results are summarized in Table~\ref{tab:post_train_ablation}. When comparing the two-stage settings, we find that replacing the final \emph{SFT (Fast)} phase with \emph{KD (Fast)} improves performance (AVG 58.66 vs. 56.28), indicating that knowledge distillation provides stronger supervision than repeated SFT alone. Extending to three-stage pipelines, introducing KD after the initial reasoning-oriented SFT further improves generalization, with the configuration \emph{SFT (Reasoning) + SFT (Fast) + KD (Fast)} achieving an average score of 58.79, slightly better than its variant with reversed order (57.90). Finally, we explore four-stage pipelines that combine both reasoning and fast-thinking KD. Among these, the sequence \emph{SFT (Reasoning) + KD (Reasoning) + SFT (Fast) + KD (Fast)} achieves the best overall performance (AVG 59.59), delivering strong gains on benchmarks such as CLUEWSC (81.25), GSM8K (73.16), and HumanEval (65.24). This suggests that a balanced alternation of SFT and KD, especially when incorporating both reasoning and fast-thinking teacher guidance, yields the most effective transfer of knowledge in the post-training phase and sets a clear direction for optimizing future training pipelines.

\section{Conclusion and Discussion}
This work introduces a comprehensive post-training pipeline designed to significantly enhance the capabilities of small language models, combining curriculum-based SFT and offline on-policy knowledge distillation. Applying this post-training pipeline, we have developed a compact yet powerful model that achieves state-of-the-art performance among models with around 1 billion parameters. All the training and deployment are conducted on Ascend hardware. Extensive experiments demonstrate that \modelname{}-KD excels in mathematical reasoning, code generation, and multilingual understanding while maintaining superior inference efficiency on Ascend edge hardware. This provides a practical and scalable solution for bridging high-performance AI and edge deployment, paving the way for advanced on-device intelligence. Future work will focus on expanding capabilities and optimizing for broader edge scenarios.

\section{Acknowledgments}
The success of this research is a collaborative effort. We are particularly grateful to the Data team for curating the training dataset, the Infrastructure team for building the robust training and inference framework, and the Evaluation team for their insightfulassessment of the model's performance.

\bibliography{ref}

\begin{thebibliography}{43}
\providecommand{\natexlab}[1]{#1}
\providecommand{\url}[1]{\texttt{#1}}
\expandafter\ifx\csname urlstyle\endcsname\relax
  \providecommand{\doi}[1]{doi: #1}\else
  \providecommand{\doi}{doi: \begingroup \urlstyle{rm}\Url}\fi

\bibitem[Achiam et~al.(2023)Achiam, Adler, Agarwal, Ahmad, Akkaya, Aleman, Almeida, Altenschmidt, Altman, Anadkat, et~al.]{achiam2023gpt}
Achiam, J., Adler, S., Agarwal, S., Ahmad, L., Akkaya, I., Aleman, F.~L., Almeida, D., Altenschmidt, J., Altman, S., Anadkat, S., et~al.
\newblock Gpt-4 technical report.
\newblock \emph{arXiv preprint arXiv:2303.08774}, 2023.

\bibitem[Agarwal et~al.(2024)Agarwal, Vieillard, Zhou, Stanczyk, Garea, Geist, and Bachem]{agarwal2024policy}
Agarwal, R., Vieillard, N., Zhou, Y., Stanczyk, P., Garea, S.~R., Geist, M., and Bachem, O.
\newblock On-policy distillation of language models: Learning from self-generated mistakes.
\newblock In \emph{The twelfth international conference on learning representations}, 2024.

\bibitem[Austin et~al.(2021)Austin, Odena, Nye, Bosma, Michalewski, Dohan, Jiang, Cai, Terry, Le, et~al.]{austin2021program}
Austin, J., Odena, A., Nye, M., Bosma, M., Michalewski, H., Dohan, D., Jiang, E., Cai, C., Terry, M., Le, Q., et~al.
\newblock Program synthesis with large language models.
\newblock \emph{arXiv preprint arXiv:2108.07732}, 2021.

\bibitem[Bai et~al.(2023)Bai, Bai, Chu, Cui, Dang, Deng, Fan, Ge, Han, Huang, et~al.]{bai2023qwen}
Bai, J., Bai, S., Chu, Y., Cui, Z., Dang, K., Deng, X., Fan, Y., Ge, W., Han, Y., Huang, F., et~al.
\newblock Qwen technical report.
\newblock \emph{arXiv preprint arXiv:2309.16609}, 2023.

\bibitem[Chen et~al.(2025)Chen, Wang, Han, Li, Li, Bi, Li, Wang, Mi, Zhu, et~al.]{chen2025pangu}
Chen, H., Wang, Y., Han, K., Li, D., Li, L., Bi, Z., Li, J., Wang, H., Mi, F., Zhu, M., et~al.
\newblock Pangu embedded: An efficient dual-system llm reasoner with metacognition.
\newblock \emph{arXiv preprint arXiv:2505.22375}, 2025.

\bibitem[Chen et~al.(2021)Chen, Tworek, Jun, Yuan, Pond{\'e}, Kaplan, Edwards, Burda, Joseph, Brockman, Ray, Puri, Krueger, Petrov, Khlaaf, Sastry, Mishkin, Chan, Gray, Ryder, Pavlov, Power, Kaiser, Bavarian, Winter, Tillet, Such, Cummings, Plappert, Chantzis, Barnes, Herbert-Voss, Guss, Nichol, Babuschkin, Balaji, Jain, Carr, Leike, Achiam, Misra, Morikawa, Radford, Knight, Brundage, Murati, Mayer, Welinder, McGrew, Amodei, McCandlish, Sutskever, and Zaremba]{Chen2021EvaluatingLL}
Chen, M., Tworek, J., Jun, H., Yuan, Q., Pond{\'e}, H., Kaplan, J., Edwards, H., Burda, Y., Joseph, N., Brockman, G., Ray, A., Puri, R., Krueger, G., Petrov, M., Khlaaf, H., Sastry, G., Mishkin, P., Chan, B., Gray, S., Ryder, N., Pavlov, M., Power, A., Kaiser, L., Bavarian, M., Winter, C., Tillet, P., Such, F.~P., Cummings, D.~W., Plappert, M., Chantzis, F., Barnes, E., Herbert-Voss, A., Guss, W.~H., Nichol, A., Babuschkin, I., Balaji, S., Jain, S., Carr, A., Leike, J., Achiam, J., Misra, V., Morikawa, E., Radford, A., Knight, M.~M., Brundage, M., Murati, M., Mayer, K., Welinder, P., McGrew, B., Amodei, D., McCandlish, S., Sutskever, I., and Zaremba, W.
\newblock Evaluating large language models trained on code.
\newblock \emph{ArXiv}, abs/2107.03374, 2021.
\newblock URL \url{https://api.semanticscholar.org/CorpusID:235755472}.

\bibitem[Cobbe et~al.(2021)Cobbe, Kosaraju, Bavarian, Chen, Jun, Kaiser, Plappert, Tworek, Hilton, Nakano, et~al.]{cobbe2021training}
Cobbe, K., Kosaraju, V., Bavarian, M., Chen, M., Jun, H., Kaiser, L., Plappert, M., Tworek, J., Hilton, J., Nakano, R., et~al.
\newblock Training verifiers to solve math word problems.
\newblock \emph{arXiv preprint arXiv:2110.14168}, 2021.

\bibitem[Du et~al.(2023)Du, Zong, and Zhang]{du2023mods}
Du, Q., Zong, C., and Zhang, J.
\newblock Mods: Model-oriented data selection for instruction tuning.
\newblock \emph{arXiv preprint arXiv:2311.15653}, 2023.

\bibitem[Dua et~al.(2019)Dua, Wang, Dasigi, Stanovsky, Singh, and Gardner]{Dua2019DROPAR}
Dua, D., Wang, Y., Dasigi, P., Stanovsky, G., Singh, S., and Gardner, M.
\newblock Drop: A reading comprehension benchmark requiring discrete reasoning over paragraphs.
\newblock In \emph{North American Chapter of the Association for Computational Linguistics}, 2019.
\newblock URL \url{https://api.semanticscholar.org/CorpusID:67855846}.

\bibitem[Dubey et~al.(2024)Dubey, Jauhri, Pandey, Kadian, Al-Dahle, Letman, Mathur, Schelten, Yang, Fan, et~al.]{dubey2024llama}
Dubey, A., Jauhri, A., Pandey, A., Kadian, A., Al-Dahle, A., Letman, A., Mathur, A., Schelten, A., Yang, A., Fan, A., et~al.
\newblock The llama 3 herd of models.
\newblock \emph{arXiv preprint arXiv:2407.21783}, 2024.

\bibitem[Fu et~al.(2024)Fu, Li, Zhao, Ma, Dutta, Zhang, Yang, Jin, and Guo]{fu2024hardware}
Fu, W., Li, S., Zhao, Y., Ma, H., Dutta, R., Zhang, X., Yang, K., Jin, Y., and Guo, X.
\newblock Hardware phi-1.5 b: A large language model encodes hardware domain specific knowledge.
\newblock In \emph{2024 29th Asia and South Pacific Design Automation Conference (ASP-DAC)}, pp.\  349--354. IEEE, 2024.

\bibitem[Groeneveld et~al.(2024)Groeneveld, Beltagy, Walsh, Bhagia, Kinney, Tafjord, Jha, Ivison, Magnusson, Wang, et~al.]{groeneveld2024olmo}
Groeneveld, D., Beltagy, I., Walsh, P., Bhagia, A., Kinney, R., Tafjord, O., Jha, A.~H., Ivison, H., Magnusson, I., Wang, Y., et~al.
\newblock Olmo: Accelerating the science of language models.
\newblock \emph{arXiv preprint arXiv:2402.00838}, 2024.

\bibitem[Guo et~al.(2024)Guo, Zhu, Yang, Xie, Dong, Zhang, Chen, Bi, Wu, Li, et~al.]{guo2024deepseek}
Guo, D., Zhu, Q., Yang, D., Xie, Z., Dong, K., Zhang, W., Chen, G., Bi, X., Wu, Y., Li, Y., et~al.
\newblock Deepseek-coder: When the large language model meets programming--the rise of code intelligence.
\newblock \emph{arXiv preprint arXiv:2401.14196}, 2024.

\bibitem[Hendrycks et~al.(2020)Hendrycks, Burns, Basart, Zou, Mazeika, Song, and Steinhardt]{hendrycks2020measuring}
Hendrycks, D., Burns, C., Basart, S., Zou, A., Mazeika, M., Song, D., and Steinhardt, J.
\newblock Measuring massive multitask language understanding.
\newblock \emph{arXiv preprint arXiv:2009.03300}, 2020.

\bibitem[Hendrycks et~al.(2021)Hendrycks, Burns, Kadavath, Arora, Basart, Tang, Song, and Steinhardt]{hendrycks2021measuring}
Hendrycks, D., Burns, C., Kadavath, S., Arora, A., Basart, S., Tang, E., Song, D., and Steinhardt, J.
\newblock Measuring mathematical problem solving with the math dataset.
\newblock \emph{arXiv preprint arXiv:2103.03874}, 2021.

\bibitem[Hinton et~al.(2015)Hinton, Vinyals, and Dean]{hinton2015distilling}
Hinton, G., Vinyals, O., and Dean, J.
\newblock Distilling the knowledge in a neural network.
\newblock \emph{arXiv preprint arXiv:1503.02531}, 2015.

\bibitem[Huang et~al.(2023)Huang, Bai, Zhu, Zhang, Zhang, Su, Liu, Lv, Zhang, Lei, Qi, Fu, Sun, and He]{Huang2023CEvalAM}
Huang, Y., Bai, Y., Zhu, Z., Zhang, J., Zhang, J., Su, T., Liu, J., Lv, C., Zhang, Y., Lei, J., Qi, F., Fu, Y., Sun, M., and He, J.
\newblock C-eval: A multi-level multi-discipline chinese evaluation suite for foundation models.
\newblock \emph{ArXiv}, abs/2305.08322, 2023.
\newblock URL \url{https://api.semanticscholar.org/CorpusID:258685666}.

\bibitem[Kashyap et~al.(2022)Kashyap, Kashyap, et~al.]{kashyap2022gpt}
Kashyap, R., Kashyap, V., et~al.
\newblock Gpt-neo for commonsense reasoning--a theoretical and practical lens.
\newblock \emph{arXiv preprint arXiv:2211.15593}, 2022.

\bibitem[Kim \& Rush(2016)Kim and Rush]{kim2016sequence}
Kim, Y. and Rush, A.~M.
\newblock Sequence-level knowledge distillation.
\newblock In \emph{Proceedings of the 2016 conference on empirical methods in natural language processing}, pp.\  1317--1327, 2016.

\bibitem[Li et~al.(2023)Li, Zhang, Koto, Yang, Zhao, Gong, Duan, and Baldwin]{Li2023cmmlu}
Li, H., Zhang, Y., Koto, F., Yang, Y., Zhao, H., Gong, Y., Duan, N., and Baldwin, T.
\newblock Cmmlu: Measuring massive multitask language understanding in chinese.
\newblock \emph{arXiv preprint arXiv:2306.09212}, 2023.

\bibitem[Li et~al.(2024)Li, Zhang, Li, Chen, Chen, Cheng, Wang, Zhou, and Xiao]{li-etal-2024-quantity}
Li, M., Zhang, Y., Li, Z., Chen, J., Chen, L., Cheng, N., Wang, J., Zhou, T., and Xiao, J.
\newblock From quantity to quality: Boosting {LLM} performance with self-guided data selection for instruction tuning.
\newblock In Duh, K., Gomez, H., and Bethard, S. (eds.), \emph{Proceedings of the 2024 Conference of the North American Chapter of the Association for Computational Linguistics: Human Language Technologies (Volume 1: Long Papers)}, pp.\  7602--7635, Mexico City, Mexico, June 2024. Association for Computational Linguistics.
\newblock \doi{10.18653/v1/2024.naacl-long.421}.
\newblock URL \url{https://aclanthology.org/2024.naacl-long.421/}.

\bibitem[Lin et~al.(2020)Lin, Wohlwend, Chen, and Lei]{lin2020autoregressive}
Lin, A., Wohlwend, J., Chen, H., and Lei, T.
\newblock Autoregressive knowledge distillation through imitation learning.
\newblock \emph{arXiv preprint arXiv:2009.07253}, 2020.

\bibitem[Liu et~al.(2024{\natexlab{a}})Liu, Feng, Xue, Wang, Wu, Lu, Zhao, Deng, Zhang, Ruan, et~al.]{liu2024deepseek}
Liu, A., Feng, B., Xue, B., Wang, B., Wu, B., Lu, C., Zhao, C., Deng, C., Zhang, C., Ruan, C., et~al.
\newblock Deepseek-v3 technical report.
\newblock \emph{arXiv preprint arXiv:2412.19437}, 2024{\natexlab{a}}.

\bibitem[Liu et~al.()Liu, Zeng, He, Jiang, and He]{liumakes}
Liu, W., Zeng, W., He, K., Jiang, Y., and He, J.
\newblock What makes good data for alignment? a comprehensive study of automatic data selection in instruction tuning.
\newblock In \emph{The Twelfth International Conference on Learning Representations}.

\bibitem[Liu et~al.(2024{\natexlab{b}})Liu, Zhao, Iandola, Lai, Tian, Fedorov, Xiong, Chang, Shi, Krishnamoorthi, et~al.]{liu2024mobilellm}
Liu, Z., Zhao, C., Iandola, F., Lai, C., Tian, Y., Fedorov, I., Xiong, Y., Chang, E., Shi, Y., Krishnamoorthi, R., et~al.
\newblock Mobilellm: Optimizing sub-billion parameter language models for on-device use cases.
\newblock In \emph{Forty-first International Conference on Machine Learning}, 2024{\natexlab{b}}.

\bibitem[Lobo et~al.(2024)Lobo, Agarwal, and Lakkaraju]{lobo2024impact}
Lobo, E., Agarwal, C., and Lakkaraju, H.
\newblock On the impact of fine-tuning on chain-of-thought reasoning.
\newblock \emph{arXiv preprint arXiv:2411.15382}, 2024.

\bibitem[Luong et~al.(2024)Luong, Zhang, Jie, Sun, Jin, and Li]{luong2024reft}
Luong, T.~Q., Zhang, X., Jie, Z., Sun, P., Jin, X., and Li, H.
\newblock Reft: Reasoning with reinforced fine-tuning.
\newblock \emph{arXiv preprint arXiv:2401.08967}, 2024.

\bibitem[Rein et~al.(2024)Rein, Hou, Stickland, Petty, Pang, Dirani, Michael, and Bowman]{rein2024gpqa}
Rein, D., Hou, B.~L., Stickland, A.~C., Petty, J., Pang, R.~Y., Dirani, J., Michael, J., and Bowman, S.~R.
\newblock Gpqa: A graduate-level google-proof q\&a benchmark.
\newblock In \emph{First Conference on Language Modeling}, 2024.

\bibitem[Ren et~al.(2024)Ren, Cao, Lin, Liu, Han, Zeng, Wan, Cai, and Sun]{ren_learning_2024}
Ren, M., Cao, B., Lin, H., Liu, C., Han, X., Zeng, K., Wan, G., Cai, X., and Sun, L.
\newblock Learning or {Self}-aligning? {Rethinking} {Instruction} {Fine}-tuning, August 2024.
\newblock URL \url{http://arxiv.org/abs/2402.18243}.
\newblock arXiv:2402.18243 [cs].

\bibitem[Ross et~al.(2011)Ross, Gordon, and Bagnell]{ross2011reduction}
Ross, S., Gordon, G., and Bagnell, D.
\newblock A reduction of imitation learning and structured prediction to no-regret online learning.
\newblock In \emph{Proceedings of the fourteenth international conference on artificial intelligence and statistics}, pp.\  627--635. JMLR Workshop and Conference Proceedings, 2011.

\bibitem[Sanh et~al.(2019)Sanh, Debut, Chaumond, and Wolf]{sanh2019distilbert}
Sanh, V., Debut, L., Chaumond, J., and Wolf, T.
\newblock Distilbert, a distilled version of bert: smaller, faster, cheaper and lighter.
\newblock \emph{arXiv preprint arXiv:1910.01108}, 2019.

\bibitem[Tang et~al.(2025)Tang, Yin, Wang, Zhou, Pan, Guo, Zhang, Rang, Liu, Zhang, et~al.]{tang2025pangu}
Tang, Y., Yin, Y., Wang, Y., Zhou, H., Pan, Y., Guo, W., Zhang, Z., Rang, M., Liu, F., Zhang, N., et~al.
\newblock Pangu ultra moe: How to train your big moe on ascend npus.
\newblock \emph{arXiv preprint arXiv:2505.04519}, 2025.

\bibitem[Team et~al.(2024)Team, Mesnard, Hardin, Dadashi, Bhupatiraju, Pathak, Sifre, Rivi{\`e}re, Kale, Love, et~al.]{team2024gemma}
Team, G., Mesnard, T., Hardin, C., Dadashi, R., Bhupatiraju, S., Pathak, S., Sifre, L., Rivi{\`e}re, M., Kale, M.~S., Love, J., et~al.
\newblock Gemma: Open models based on gemini research and technology.
\newblock \emph{arXiv preprint arXiv:2403.08295}, 2024.

\bibitem[Team et~al.(2025)Team, Xiao, Li, Han, Bai, Cai, Chen, Chen, Cong, Cui, et~al.]{team2025minicpm4}
Team, M., Xiao, C., Li, Y., Han, X., Bai, Y., Cai, J., Chen, H., Chen, W., Cong, X., Cui, G., et~al.
\newblock Minicpm4: Ultra-efficient llms on end devices.
\newblock \emph{arXiv preprint arXiv:2506.07900}, 2025.

\bibitem[Touvron et~al.(2023)Touvron, Lavril, Izacard, Martinet, Lachaux, Lacroix, Rozi{\`e}re, Goyal, Hambro, Azhar, et~al.]{touvron2023llama}
Touvron, H., Lavril, T., Izacard, G., Martinet, X., Lachaux, M.-A., Lacroix, T., Rozi{\`e}re, B., Goyal, N., Hambro, E., Azhar, F., et~al.
\newblock Llama: Open and efficient foundation language models.
\newblock \emph{arXiv preprint arXiv:2302.13971}, 2023.

\bibitem[Xu et~al.(2020)Xu, Hu, Zhang, Li, Cao, Li, Xu, Sun, Yu, Yu, et~al.]{xu2020clue}
Xu, L., Hu, H., Zhang, X., Li, L., Cao, C., Li, Y., Xu, Y., Sun, K., Yu, D., Yu, C., et~al.
\newblock Clue: A chinese language understanding evaluation benchmark.
\newblock \emph{arXiv preprint arXiv:2004.05986}, 2020.

\bibitem[Yang et~al.(2025)Yang, Li, Yang, Zhang, Hui, Zheng, Yu, Gao, Huang, Lv, et~al.]{yang2025qwen3}
Yang, A., Li, A., Yang, B., Zhang, B., Hui, B., Zheng, B., Yu, B., Gao, C., Huang, C., Lv, C., et~al.
\newblock Qwen3 technical report.
\newblock \emph{arXiv preprint arXiv:2505.09388}, 2025.

\bibitem[Yang et~al.(2024)Yang, Yang, Zhang, Hui, Zheng, Yu, Li, Liu, Huang, Dong, Wei, Lin, Yang, Tu, Zhang, Yang, Yang, Zhou, Lin, Dang, Lu, Bao, Yang, Yu, Li, Xue, Zhang, Zhu, Men, Lin, Li, Xia, Ren, Ren, Fan, Su, Zhang, Wan, Liu, Cui, Zhang, Qiu, Quan, and Wang]{Yang2024Qwen25TR}
Yang, Q.~A., Yang, B., Zhang, B., Hui, B., Zheng, B., Yu, B., Li, C., Liu, D., Huang, F., Dong, G., Wei, H., Lin, H., Yang, J., Tu, J., Zhang, J., Yang, J., Yang, J., Zhou, J., Lin, J., Dang, K., Lu, K., Bao, K., Yang, K., Yu, L., Li, M., Xue, M., Zhang, P., Zhu, Q., Men, R., Lin, R., Li, T., Xia, T., Ren, X., Ren, X., Fan, Y., Su, Y., Zhang, Y.-C., Wan, Y., Liu, Y., Cui, Z., Zhang, Z., Qiu, Z., Quan, S., and Wang, Z.
\newblock Qwen2.5 technical report.
\newblock \emph{ArXiv}, abs/2412.15115, 2024.
\newblock URL \url{https://api.semanticscholar.org/CorpusID:274859421}.

\bibitem[Yin et~al.(2024)Yin, Wu, Wang, Wang, Guo, Wang, Liu, Tang, Lian, and Chen]{yin2024entropylawstorydata}
Yin, M., Wu, C., Wang, Y., Wang, H., Guo, W., Wang, Y., Liu, Y., Tang, R., Lian, D., and Chen, E.
\newblock Entropy law: The story behind data compression and llm performance, 2024.
\newblock URL \url{https://arxiv.org/abs/2407.06645}.

\bibitem[Zhang et~al.(2024)Zhang, Zeng, Wang, and Lu]{zhang2024tinyllama}
Zhang, P., Zeng, G., Wang, T., and Lu, W.
\newblock Tinyllama: An open-source small language model.
\newblock \emph{arXiv preprint arXiv:2401.02385}, 2024.

\bibitem[Zhang et~al.(2022)Zhang, Roller, Goyal, Artetxe, Chen, Chen, Dewan, Diab, Li, Lin, et~al.]{zhang2022opt}
Zhang, S., Roller, S., Goyal, N., Artetxe, M., Chen, M., Chen, S., Dewan, C., Diab, M., Li, X., Lin, X.~V., et~al.
\newblock Opt: Open pre-trained transformer language models.
\newblock \emph{arXiv preprint arXiv:2205.01068}, 2022.

\bibitem[Zhou et~al.(2023{\natexlab{a}})Zhou, Liu, Xu, Iyer, Sun, Mao, Ma, Efrat, Yu, Yu, Zhang, Ghosh, Lewis, Zettlemoyer, and Levy]{zhou_lima_2023}
Zhou, C., Liu, P., Xu, P., Iyer, S., Sun, J., Mao, Y., Ma, X., Efrat, A., Yu, P., Yu, L., Zhang, S., Ghosh, G., Lewis, M., Zettlemoyer, L., and Levy, O.
\newblock {LIMA}: {Less} {Is} {More} for {Alignment}, May 2023{\natexlab{a}}.
\newblock URL \url{http://arxiv.org/abs/2305.11206}.
\newblock arXiv:2305.11206 [cs].

\bibitem[Zhou et~al.(2023{\natexlab{b}})Zhou, Lu, Mishra, Brahma, Basu, Luan, Zhou, and Hou]{zhou2023instruction}
Zhou, J., Lu, T., Mishra, S., Brahma, S., Basu, S., Luan, Y., Zhou, D., and Hou, L.
\newblock Instruction-following evaluation for large language models.
\newblock \emph{arXiv preprint arXiv:2311.07911}, 2023{\natexlab{b}}.

\end{thebibliography}
\bibliographystyle{icml2025}

\end{document}